\documentclass{article}

\usepackage{PRIMEarxiv}

\usepackage[utf8]{inputenc} 
\usepackage[T1]{fontenc}    
\usepackage{hyperref}       
\usepackage{url}            
\usepackage{booktabs}       
\usepackage{amsfonts}       
\usepackage{nicefrac}       
\usepackage{microtype}      
\usepackage{lipsum}
\usepackage{fancyhdr}       
\usepackage{graphicx}       
\usepackage{amsmath,amsthm}
\usepackage{tabularx}
\usepackage{amsfonts}
\graphicspath{{media/}}     

\title{\textbf{Sup3r}: A Semi-Supervised Algorithm for increasing Sparsity, Stability, and Separability in Hierarchy Of Time-Surfaces architectures.} 

\author{
  Marco Rasetto \\
  Department of Bioengineering \\
  University of Pittsburgh, Center of Neural Basis of Cognition  \\
  Pittsburgh, USA\\
  \texttt{marcorasetto@pitt.edu} \\
  \And 
  Himanshu Akolkar \\
  Department of Neurobiology \\
  University of Pittsburgh,\\
  Pittsburgh, USA\\
  \texttt{akolkar@pitt.edu} \\
  \And
  Ryad Benosman \\
  Meta Reality Labs,\\
  New York, USA\\
}

\begin{document}
\maketitle

\begin{abstract}
The Hierarchy Of Time-Surfaces (HOTS) algorithm, a neuromorphic approach for feature extraction from event data, presents promising capabilities but faces challenges in accuracy and compatibility with neuromorphic hardware. In this paper, we introduce Sup3r, a Semi-Supervised algorithm aimed at addressing these challenges. Sup3r enhances sparsity, stability, and separability in the HOTS networks. It enables end-to-end online training of HOTS networks replacing external classifiers, by leveraging semi-supervised learning. Sup3r learns class-informative patterns, mitigates confounding features, and reduces the number of processed events. Moreover, Sup3r facilitates continual and incremental learning, allowing adaptation to data distribution shifts and learning new tasks without forgetting. Preliminary results on N-MNIST demonstrate that Sup3r achieves comparable accuracy to similarly sized Artificial Neural Networks trained with back-propagation. This work showcases the potential of Sup3r to advance the capabilities of HOTS networks, offering a promising avenue for neuromorphic algorithms in real-world applications.\end{abstract}


\section*{Introduction}
Hierarchy Of Time-Surfaces is a neuromorphic algorithm used to extract features from patterns of events \cite{lagorce2016hots}. This is possible thanks to a type of representation called time-surface or time vector, where events are interpolated by exponential decay kernels and collected to represent relative time differences between the activation of units in the network. Time surfaces are one of the most common representations in the neuromorphic field since they allow to interface event data with traditional machine learning and computer vision algorithms \cite{gallego2020event,gehrig2019end}. In HOTS, time surfaces are clustered together using algorithms like $k$-means to extract common activity patterns, and layers of units are built by considering each centroid as a neuron that can emit a new event when an input time surface is assigned to it. For this reason, HOTS shares many points in common with bag-of-words or bag-of-features algorithms\cite{csurka2004visual}. For instance, HOTS requires an external classifier on histograms of features to classify information. Similarly to bag-of-words algorithms, HOTS classifiers are histograms that accumulate features over a given temporal window to produce an input vector to traditional machine learning algorithms like Support Vector Machines and Multi-Layer Perceptrons\cite{HOTSGestures,rasetto2021event,lagorce2016hots, rasetto2023building}. This approach limits compatibility with neuromorphic hardware and can nullify latency and energy efficiency advantages that are found in neuromorphic systems. Compared to Spiking Neural Networks (SNNs) trained with backpropagation through time, HOTS lags in accuracy \cite{sironi2018hats, grimaldi2023robust, shrestha2018slayer,lagorce2016hots}. However, feature engineering and model tweaking can often bridge the gap on selected tasks \cite{grimaldi2023robust, sironi2018hats}. This is not surprising as it is known that clustering algorithms can reach comparable accuracy to small Convolutional Neural Networks (CNNs) when carefully tuned \cite{Coates2012}.
Moreover, clustering-based algorithms like HOTS are still worth investigating because of properties that might be inherited from clustering, like better explainability \cite{oyelade2016clustering, jaeger2023cluster} compared to more complex SNNs and continual and incremental learning\cite{convergencekmeans,pratama2021unsupervised,lamers2023clustering}. This work presents Sup3r, a Semi-Supervised algorithm for increasing Sparsity Stability and Separability in the Hierarchy of Time-Surfaces. This algorithm can train end-to-end HOTS networks online, ditching the classifier for a Deep Neural Network architecture where every last unit encodes for a different label. We show that Sup3r can learn class-informative patterns and rejects events from confounding features, reducing the number of processed events. Moreover, Sup3r allows the HOTS network to adapt to distribution shifts in the data (continual learning) and learn new tasks without forgetting (incremental learning). Finally, preliminary results on N-MNIST show that Sup3r can reach comparable accuracy to a similarly sized Artificial Neural Network (ANN) trained with back-propagation.

\section{Time Surfaces and Time Vectors}

HOTS originally introduced the concept of time surfaces \cite{lagorce2016hots}. These descriptors encode relative timings between events of units within a defined neighborhood and were historically used on events generated by event-based cameras \cite{SDVS}. This type of camera does not output frames but an asynchronous stream of events generated by individual pixels detecting brightness changes. Thus, in the context of vision, an event $i$ can be described as the tuple:

\begin{equation} \label{eq_ch3:event}
ev_i = (x_i,y_i,p_i,t_i)
\end{equation}

where $t_i$ is the timestamp of the event, $x_i$ and $y_i$ are the horizontal and vertical coordinates of the pixels generating the event, and $p_i\in [0,1]$ is the polarity of the event, which is 0 when the event signifies a decrease in brightness for the given pixel and 1 when it signifies an increase.
To generalize to different modalities and uses, we can redefine events as:

\begin{equation} \label{eq_ch3:event_general}
ev_i = (\mathbf{x_i},p_i,t_i)
\end{equation}

where $\mathbf{x}$ is now a vector (in \textbf{bold}) representing any possible space coordinates depending on the event-based sensor type. In the case of events generated by event-based cameras  $\mathbf{x} = (x,y)$ \cite{lagorce2016hots}, whereas in the case of neuromorphic cochleas $\mathbf{x} = ch$ representing the channel index \cite{rasetto2021event}. To generalize, we consider  $p_i \in \mathbb{N}$  as we now use it to index events produced by multiple units at the same spatial location \cite{lagorce2016hots, rasetto2021event}. 
Time surfaces are built by centering a squared window $R_i$ of lateral size $l$ around every $\mathbf{x_i}$ and collecting the last event emitted by neighboring pixels in a time context $T_i$:

\begin{equation}\label{eq_ch3:time_context}
    \mathbf{T_i} = \max_{j\leq i}\{t_j | \mathbf{x_j} \in R_i \} 
\end{equation}

Finally, time surfaces are built by applying an exponential decay kernel on the time context:

\begin{equation}\label{eq_ch3:time_surface}
    \mathbf{ts_i} = e^{-\frac{(t_i - \mathbf{T_i})}{\tau}}
\end{equation}

Time vectors $\mathbf{tv_i}$ are a more general definition of time surface where the number of dimensions is one or more than two \cite{rasetto2021event}. 
In HOTS, time surfaces are clustered with clustering algorithms like $k$-means. Whenever a time surface is assigned to a centroid, it produces an event. These events can be used to generate new time surfaces that can be clustered with longer taus to integrate and extract more complex patterns of activities \cite{lagorce2016hots}. Events outputted by a layer $k$ are so defined: 

\begin{equation} \label{eq_ch3:output_events}
ev_i^{k+1} = (\mathbf{x_i}^{k+1},p_i^{k+1},t_i^{k+1})
\end{equation}

where $\mathbf{x_i}^{k+1}=\mathbf{x_i}^{k}$, $t_i^{k+1}=t_i^{k}$, and $p_i^{k+1}=f_i^{k}$ with $f$ being the "firing" feature/centroid assigned to the input time surface $ts_i^k$.

\section{A hierarchy of time surfaces and time vectors}
\label{sec:ch3_hierarchy}

A fundamental principle behind HOTS networks is the increase of spatial neighborhood $R^k$ and $\tau^k$ at every layer to extract longer and more complex spatial and temporal patterns. In this work, we use the same principle, with the difference of the last layer $K$ (where $K$ denotes the last layer of a network with layer index $k$) where $R^K$ is always including all coordinates $\mathbf{x}$. This is somewhat different from traditional HOTS implementation, where it is often the external classifier that integrates all the possible spatial coordinates.

\section{Sparsity, Separability, and Stability in spiking networks}

Sup3r aims to remove external classifiers from HOTS by using the network's last layer as a classifier for a pattern recognition problem. In that case, we expect the number of clusters to equal the number of classes in the problem and for each cluster to generate events only when the correct class is presented to the network. In this context, the classifier will have to limit the firing to only one unit (high \textbf{Sparsity}) for the whole duration of the stimuli (high \textbf{Stability}). In the context of clustering, this also means that classes assigned to different centroids must be well \textbf{Separated}.
Conveniently, we can represent these three concepts using a time vector called feedback-time-vector $\mathbf{ftv}^K$, built on the output events $ev_i^{K+1}$ of our classifier.  Rather than pulling the last event in a given neighborhood $R$ (eq. \ref{eq_ch3:time_context}),  we pull the last event of every polarity in the last layer:

\begin{equation}\label{eq_ch3:last_time_context}
    \mathbf{FT_i^K} = \max_{j\leq i}\{t_j^{K+1} | \mathbf{p_j^{K+1}}\} 
\end{equation}

We use $\mathbf{ftv}^k$ to differentiate feedback time vectors from standard time vectors $\mathbf{tv}^K$ or time surfaces $\mathbf{ts^K}$ built on the input events $ev_i^K$. We then use it to build the descriptor $S$:
\begin{equation}
    S^{K}_i = G(\mathbf{ftv}^K_i[f]-\sum\limits_{l\neq f}^{N^K}\frac{\mathbf{ftv}^K_i[l]}{K-1})
    \label{eq:S}
\end{equation}
Where $f$ is the index of the feedback-time-vector element corresponding to the ''firing'' centroid (last event) that triggered its generation, and $N^K$ is the number of clusters for layer $K$. $G$ is a sign function that is positive if the ''firing'' centroid corresponds to the correct class label, negative if otherwise: 
\begin{equation}
    G =
    \begin{cases}
        + & \text{for } f=cc \\
        - & \text{for } f \neq cc
    \end{cases}
    \label{eq:G}
\end{equation}
where $cc$ is the index of the correct class.
To see how this descriptor relates to Sparsity, Stability, and Separability, let us look at the limits for $S$. If the classifier has a stable, sparse, and correct response (only one unit active, corresponding to the correct class index $cl$), $S$ will tend to 1:
\begin{equation}
    S^{K}_i = +(1 - \sum\limits_{l\neq f}^{N^K}\frac{0}{K-1}) = 1
    \label{eq:S=1}
\end{equation}

Conversely, if only one unit is active, but it corresponds to the wrong class label, G will be negative:
\begin{equation}
    S^{K}_i = -(1 - \sum\limits_{l\neq f}^{N^K}\frac{0}{K-1}) = -1
    \label{eq:S=-1}
\end{equation}
indicating a fully sparse and stable but wrong response, meaning the classifier does not correctly separate the classes.

In case the activity of the classifier is dense, meaning that all clusters are firing events at approximately the same time, the feedback time vectors will become fully populated by ones, making $S=0$:
\begin{equation}
    S^{K}_i = G(1 - \sum\limits_{l\neq f}^{N^K}\frac{1}{K-1}) = G(1 - \frac{K-1}{K-1}) = G(0)= 0
    \label{eq:S=0}
\end{equation}

Thus, $S$ amplitude $\in[0,1]$ represents the degree of sparsity and stability of the last layer activation, where 1 denotes a fully sparse and stable response, and 0 indicates a dense and unstable activity. We can extend this descriptor to every layer of the network $k$, to characterize the degree of Sparsity, Stability, and Separability of all the centroids in multi-layer networks. The only difference with layer K is that intermediate layers will output events with spatial coordinates since they are not integrating over all coordinates $\mathbf{x}$ (sec. \ref{sec:ch3_hierarchy}). Therefore, the feedback time context $FT_i^k$ will be defined as such:

\begin{equation}\label{eq_ch3:feedback_time_context}
    \mathbf{FT_i^k} = \max_{j\leq i}\{t_j^{k+1} | \mathbf{x_i^{k+1}}, \mathbf{p_j^{k+1}}\} 
\end{equation}

meaning that $\mathbf{FT_i^k}$ pulls events with different polarities $\mathbf{p_j^{k+1}}$ and the same spatial coordinates $\mathbf{x_i^{k+1}}$.

\section{Learning class-relevant features}

Since we want to develop an online and local learning rule, we cannot resort to backpropagation to maximize $S$, especially if we consider that HOTS neurons are not differentiable. Therefore, we resort to a different approach, focusing on using a modified online $k$-means learning rule. The learning rule we propose is described by the following equation:

\begin{equation}
    \Delta \mathbf{c}^{k}_f = \alpha \Delta S^{k+1}_i \mathbf{q}^k_{i,f} + \beta S^{k+1}_i \mathbf{q}^k_{i,f}
    \label{eq:learningruleK}
\end{equation}

where $\alpha$ and$\beta$ are learning rates, $\mathbf{c}^k_f$ represents the position of the ''firing'' centroid, assigned to a novel time surface $\mathbf{ts}^k _i$ and $\mathbf{q}^k_{i,f} = \mathbf{ts}^{k}_i - \mathbf{c}^k_f$. If we  assume $S^k_i, \Delta S^k_i = 1$ and $\eta=\alpha+\beta$ the equation takes the form of online $k$-means \cite{convergencekmeans}: 

\begin{equation}
    \Delta \mathbf{c}^k_f = \eta \, \mathbf{q}^k_{i,f} = \eta \, (\mathbf{ts}^k _i - \mathbf{c}^k_f)
    \label{eq:onlinekmeans}
\end{equation}

As shown (eq: \ref{eq:learningruleK}), the proposed learning rule comprises two terms that weight the online $k$-means equation (eq: \ref{eq:onlinekmeans}) by $\Delta S$ and $S$.  The first term allows centroids to only move in the direction of a set of features when they are associated with an increase in $S$. Conversely, centroids will move away from features that decrease $S$ (such as features of the wrong class or class-unrelated features), ensuring that only class-relevant features are learned. As $S \rightarrow 1$, $\Delta S \rightarrow 0$  that makes the first term of equation \ref{eq:learningruleK} tend to 0. This ensures the learning rate will decrease as we approach a maximum, but it can cause the centroids never to reach the center of mass of clusters. For this reason, the second term of the learning rule is weighted by $S$ but with a different learning rate $\beta<\alpha$ to ensure it is not adding instability to local maxima. 

\section{Individual thresholds}
One problem arising from the presented learning rule is the choice of the decision boundaries for the cluster assignments. In $k$-means, the decision boundaries are defined as the equidistant points between two adjacent clusters, similar to the borders of a Voronoi diagram. Inconveniently, the learning rule presented with equation \ref{eq:learningruleK} is not compatible with this type of decision boundary. The reason is that this type of segmentation will always assign every sample to a centroid, making it impossible to learn only class-relevant features. This means that when some centroids are pushed to more relevant features, the less relevant features will be assigned to other nearby clusters, reducing their specificity and ultimately lowering classification accuracy. To avoid this scenario, we introduce individual thresholds $th^k_n$, where n is the cluster index for layer $k$. We initially assign every sample to the closest cluster as done in $k$-means, but now, every cluster will generate a new event only when the sample distance to the cluster is less than $th^k_n$. When this happens, we tune individual thresholds with the equation: 

\begin{equation}
    \Delta th^{k}_f = \gamma \Delta S^{k+1}_i  e^{-\frac{\lVert \mathbf{q}^k_{i,f}\rVert_2}{d^k}}  + \delta S^{k+1}_i e^{-\frac{\lVert \mathbf{q}^k_{i,f}\rVert_2}{d^k}}
    \label{eq:learningruleth}
\end{equation}

where $d$ is a parameter used to control how close to the border samples need to be to have an effect on the threshold update. We also update nearby thresholds with a competitive rule to reduce the number of events produced by the network and threshold overlap between adjacent clusters. If a sample falls within the decision boundaries of other ''non-firing'' clusters, represented by the index $nf$, and $\Delta S$ and $S$ are positive, we update their thresholds with the following equation:

\begin{equation}
    \Delta th^{k}_{nf} = - \gamma \Delta S^{k+1}_i  e^{-\frac{\lVert \mathbf{q}^k_{i,nf}\rVert_2}{d^k}}  - \delta S^{k+1}_i e^{-\frac{\lVert \mathbf{q}^k_{i,nf}\rVert_2}{d^k}}
    \label{eq:learningruleth_nonf}
\end{equation}

Similarly to equation \ref{eq:learningruleK}, both rule presents two terms: one weighted by $\Delta S^{k+1}_i$ and one weighted by $S^{k+1}_i$. The first term changes the decision boundaries to include samples associated with an increase of $S$ and reject samples associated with its decrease.  In the situation where the network is consistently wrong or consistently right, this first term will tend to 0, requiring a second term weighted by $S^{k+1}_i$ to increase network firing rate and stability for correct classes and reduce them for incorrect ones. As noted for equation \ref{eq:learningruleK}, $\delta<\gamma$ to ensure the stability of the learning rule.  

\section{Network Initialization}

Similarly to $k$-means, centroid initialization is an important step to ensure good results \cite{gul2023big}. 
Random initialization might put the centroids too far away from the feature spaces or regions where the sum of updates from the learning rule might push centroids even further from any meaningful features. However, we are not interested in state-of-the-art initialization techniques since they might obfuscate the actual performance results of Sup3r. Therefore, we initialize centroids on the average time surfaces from a small batch of data and add noise drawn from a uniform distribution:

\begin{equation}\label{eq:initialization}
    \mathbf{c_j^k} = (1-\zeta) \, \overline{\mathbf{ts^k}} + \zeta \, \mu_{ts} \, \mathbf{U_j^k}
\end{equation}

Were $\mathbf{U}$ represents a matrix with the same shape as $\mathbf{c_j^k}$ and composed of random values pulled from a uniform distribution. Since $\mathbf{ts_i}$ with small $\tau$ are mostly 0 while $\mathbf{ts_i}$ with high $\tau$ are mostly 1, we scale $\mathbf{U}$ by the average of the elements in $\overline{\mathbf{ts^k}}$, defined as $\mu_{ts}$. This approach is extremely simple. In the future, more complex approaches could be explored to ensure the convergence of most features and boost learning speed and accuracy.

\section*{Results}

\subsection{Synthetic benchmark}
\label{sec:synth_class}

\begin{figure}
    \centering
    \includegraphics[width=0.8\textwidth]{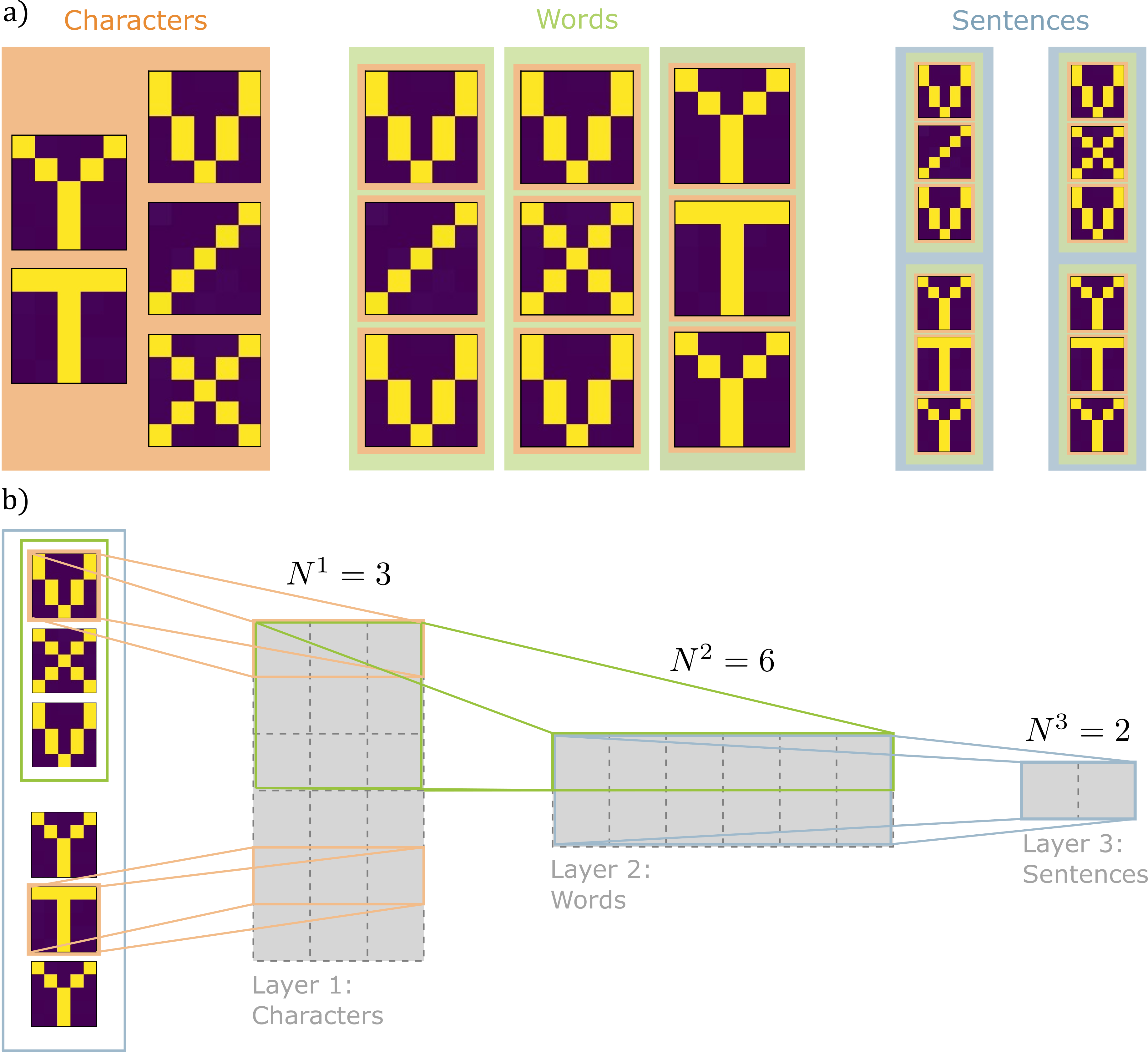}
    \caption{The visual classification benchmark to test Sup3r on a small 3-layered network. a) The test is composed of a ''sentence'' (blue) classification task, where two different sentences are each composed of two among three different ''words'' (green), which are composed of three of five different ''characters'' (orange). The only difference between the two sentences consists of two characters, ''x'' and ''/'', which are contained in the top word. To generate spiking data, sentences are written by a lattice of 5x30 Poisson neurons firing at a high firing rate for the yellow pixels and a low firing rate for the blue pixels. b) The convolutional HOTS network used to solve the task. The gray rectangles represent the output dimensionality of each layer. $N$ is the number of clusters responding to the six different characters in a sentence (Layer 1), to the two words in a sentence (Layer 2), and finally to the entire sentence (Layer 3).}
    \label{fig:synth_bench}
\end{figure}

Hierarchy of Time-Surfaces uses unsupervised clustering algorithms such as online \textit{k-}means to extract recurring features in the data. While this approach can be effective for pre-training on large quantities of unlabeled data, it presents several drawbacks compared to state-of-the-art neuromorphic networks that use approximate gradient methods based on Back Propagation Through Time. One problem is the need for an external non-neuromorphic classifier, which we outlined in the introduction. A second problem is the loss of accuracy given by learning non-class-relevant features. To best represent this problem, we design a tailored neuromorphic synthetic benchmark. We propose a visual pattern classification task requiring a three-layered network  (fig. \ref{fig:synth_bench}). The proposed task is to classify two ''sentences'' (''v/v yty'' and ''vxv yty'') composed of two ''words'' each made of three ''characters'' drawn by lattice of 5x5 Poisson spiking neurons, for a total of 5x30 neurons (as can be seen in fig. \ref{fig:synth_bench}a). Yellow pixels correspond to ''high'' firing rate while blue pixels correspond to ''low'' firing rate. We use Poisson neurons and background firing rate to introduce variability per every sample. The training set consists of 1000 different recordings (500 for each sentence) of 10ms. This visual pattern is then classified by a three-layered convolutional HOTS neural network (fig. \ref{fig:synth_bench}b) where the first layer ($N^1=3$) responds to individual characters, the second layer ($N^2 = 6$) responds to single words, and the last layer ($N^3 = 2$) classifies the entire sentence.  It should be noted that the number of HOTS centroids for each layer  ($N_i$) does not correspond to the actual number of clusters in the data. In the first layer, the number of clusters is five (all the possible characters: ''v, /, x, y, t''), while in the second layer, this number is three (all the possible words in the dataset: ''v/v, vxv, yty''). This choice is motivated by the need to represent real-world scenarios in which the number of clusters is unknown. This can be especially problematic when the number of centroids is less than the actual number of clusters in the data, as similar clusters representing two features of distinct classes might be assigned a single centroid. This is the case of layer 1, making it likely for the same centroid to represent similar characters like ''x'' and ''/''. These two characters are the only difference in the two sentences in the dataset (fig \ref{fig:synth_bench}a), making them class-relevant features. Failure to assign separate centroids to  ''x'' and ''/'' means that layer 1 will respond similarly to the two sentences, causing the other network layers to fail to separate the two sentences.

\subsubsection{Classification Results}

\begin{figure}
    \centering
    \includegraphics[width=1\textwidth]{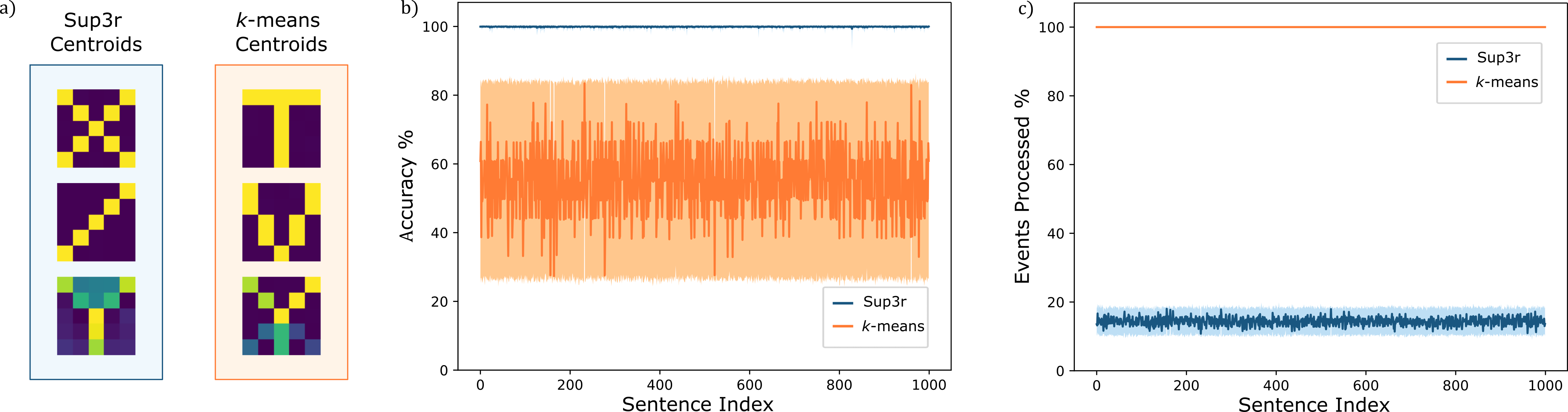}
    \caption{Neuromorphic benchmark results, comparing Sup3r against $k$-means for feature extraction. a) Layer 1 features extracted with $k$-means (light blue) and Sup3r (orange). $k$-means fails to extract class-relevant features, while Sup3r can correctly extract   ''x''  ''/'', leaving one centroid unused. b) Consequently, $k$-means fails the classification task, performing close to the chance level ($\approx$55.36\% accuracy, calculated as the percentage of events assigned to the correct class), while Sup3r can solve the task with $\approx$99.92\% accuracy. Since Sup3r rejects events from non-class-relevant characters, only $\approx$14.33\% of events are propagated in the network (calculated as the ratio between the number of the last layer output events divided by the number of the first layer input events). Conversely, the k-means network outputs the same number of input events.}
    \label{fig:bench-res}
\end{figure}

We show the classification results of a separate test set of 1000 recordings (500 recordings for each sentence). We test Sup3r against a comparable network with \textit{k}-means features. Figure \ref{fig:bench-res} shows the features from layer 1 for both architectures. As expected, Sup3r can leverage Separability Sparsity and Stability from higher layers to extract ''x'' and ''/''   (fig. \ref{fig:bench-res}a) without utilizing the third centroid. Oppositely, \textit{k}-means must assign every centroid to a group of samples and minimize their variance, an optimization technique that forces similar features to share a centroid when the number of centroids is below the number of actual clusters in the dataset. In figure \ref{fig:bench-res}b, we define the network accuracy by calculating the percentage of events assigned to the right sentence in a single recording. The \textit{k}-means algorithm causes HOTS to fail to recognize the two sentences, resulting in $\approx$55.36\% accuracy (close to the chance level). Conversely, Sup3r reaches  $\approx$99.92\% accuracy while using only two centroids. Since Sup3r can reject any event produced by the non-class-relevant features, it outputs only  $\approx$14.33\% of the input events (fig. \ref{fig:bench-res}c), potentially reducing energy consumption in hardware implementations. To train the network we used the following parameters: $\alpha=1 \cdot 10^{-4}$, $\beta = 1 \cdot 10^{-5}$, $\gamma=1 \cdot 10^{-4}$, $\delta = 5\cdot 10^{-6}$, $\tau_1 = 1s$,$\tau_2 = 1ms$, $\tau_3 = 1ms$ and \textbf{ftv} taus $f\tau_2=100ms$ and $f\tau_3=10ms$. We also report the results in table \ref{tab:class}.

\begin{table}
\small
\caption{Classification results (Mean on 10 networks trained and tested on independently generated synthetic datasets)}
\centering
\begin{tabular}{l|ll}

            & Accuracy& Processed events\\ \hline
Sup3r Network& 99.92\%& 14.33\%\\
 $k$-means& 55.53\%&100\%\\\end{tabular}
\label{tab:class}
\end{table}

\subsubsection{Continual Learning}

\begin{figure}
    \centering
    \includegraphics[scale=0.8]{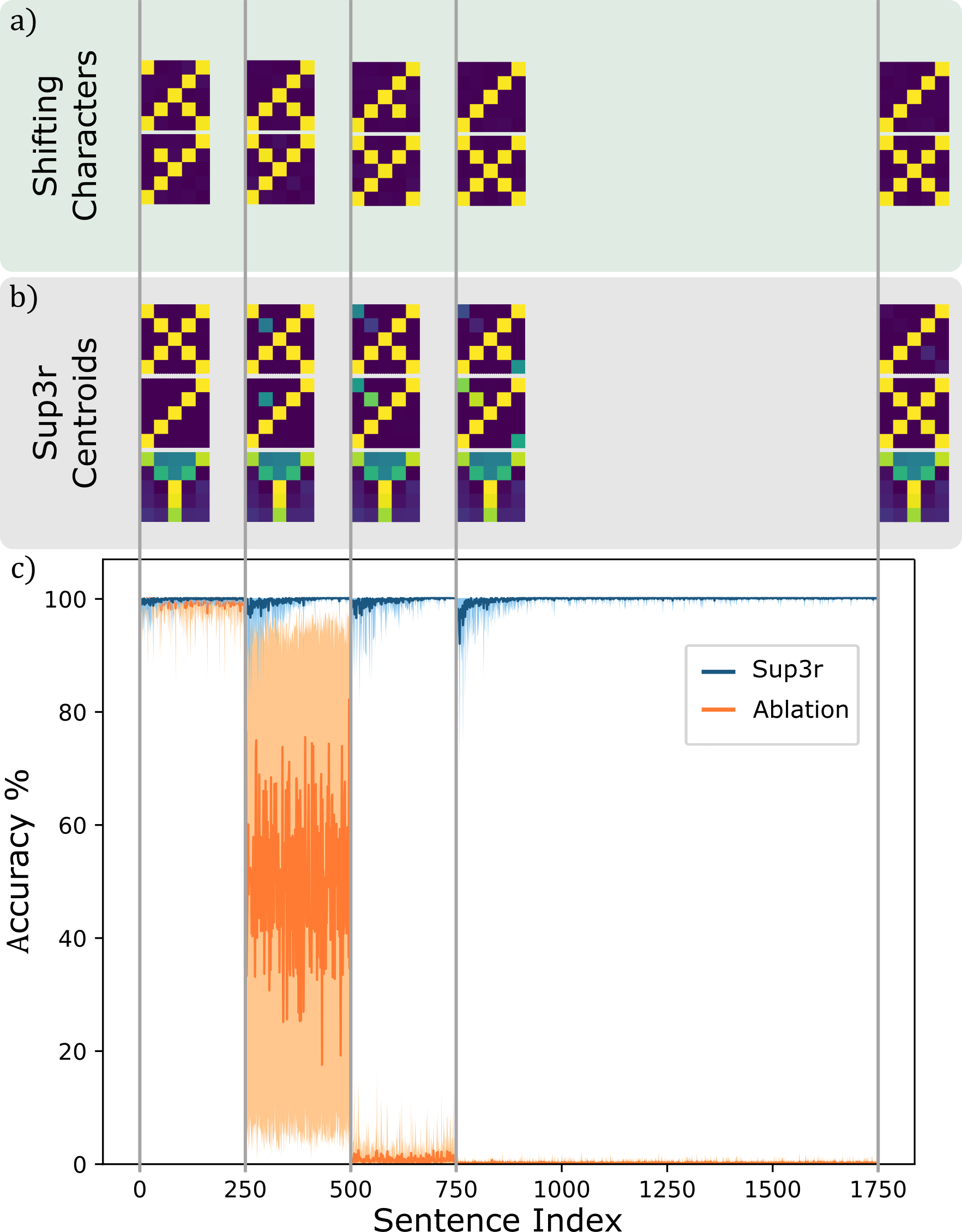}
    \caption{Continual learning test with Sup3r. In this test, we perform a class shift in the two sentences by subtracting pixels from ''x'' and adding them to ''/'' every 250 sentences (a), with the final shift lasting 1000 sentences. Sup3r centroids slowly adapt to the shift (b). In (c), Sup3r network accuracy remains high and returns to >99\%, while the same network without unsupervised learning fails the task (Ablation). This test is performed with ten independently trained networks and randomized test sets. Accuracy is reported as the average across runs in dark colors with min and max values in light colors. }
    \label{fig:continual}
\end{figure}

Sup3r learning signal $S_i^K$ is composed of a local signal (coming only from the next layer ) and a global signal $G$. It is important to note that the local signal is entirely unsupervised, as there is no need to know the correct label to calculate it (eq. \ref{eq:S}). This opens the possibility of keeping the unsupervised learning running while inferring to adapt to potential shifts in the class distributions. To test this feature, we design a test set of 1750 recordings by shifting ''x'' and ''/'' characters in the test sentences. To shift the characters, we subtract pixels from “x” to add them to ''/'' incrementally, with the final shift causing the two letters to swap entirely (fig. \ref{fig:continual}a). Every shift lasts for 250 recordings, with the last shift lasting for 1000 to ensure convergence and stability. Looking at the Sup3r centroids, we can see them adapting to the shift so that ''x'' and ''/'' become swapped (fig. \ref{fig:continual}b). To evaluate Sup3r accuracy, in figure \ref{fig:continual}c, we test the network against an ablated network with no unsupervised learning. As expected, Sup3r adapts to distribution shifts while maintaining the original accuracy (>99\%), while the ablated network fails the task, ending with 0\% accuracy. This final test was performed with 10 independently trained networks with 10 distinct test sets. To train the network, we used the same parameters we used for the classification results. 

\subsubsection{Incremental Learning}

Our hypothesis is that Sup3r ability to extract sparse class-relevant features makes it a natural candidate for Incremental Learning. Since we are extracting a set of sparse class-relevant features, Sup3r might learn new tasks simply by adding new centroids and fixing the previous ones to avoid catastrophic forgetting.

To test this hypothesis, we design a new task (task 2) in which a Sup3r network has to learn two novel sentences, "vpv yty" and "vov yty", after being trained on the previous classification task (task 1) in section \ref{sec:synth_class}. 
We then tested the same network on task 1 after retraining on task 2 to demonstrate that it did not suffer catastrophic forgetting. We show the results in Table \ref{tab:Incremental}.

\begin{table}
\small
\caption{Incremental Learning results (Mean on 10 networks trained and tested on independently generated synthetic datasets)}
\centering
\begin{tabular}{l|lll}

            & Task 1 & Task 2 &Task 1 after Task 2\\ \hline
Sup3r Network& 99.92\%& 99.99\%&99.96\%\\\end{tabular}
\label{tab:Incremental}
\end{table}
The results show that the network can train on Task 2 and preserve the accuracy previously reached on Task 1. To train the network, we used the same parameters we used for the classification results.

\section{Sup3r against backpropagation}

We present preliminary results to show how Sup3r compares against backpropagation. We use the N-MNIST dataset  \cite{NMNIST} a neuromorphic version of the popular MNIST dataset \cite{deng2012mnist}. This allows us to compare a Sup3r-HOTS network with a traditional ANN. The Sup3r HOTS network has a single hidden layer with 32 centroids and spatial neighborhood  $R^1=9$. The ANN has the same architecture with a single convolutional hidden layer with 32 units with sigmoid activation and filter size 9x9. We train the ANN with stochastic gradient descent and a learning rate of 0.05. Other Sup3r parameters where $\tau_1=100ms$, $\tau_2=1ms$ , $\alpha=5 \cdot 10^{-3}$, $\beta = 5\cdot 10^{-6}$, $\gamma=1 \cdot 10^{-4}$, $\delta = 1\cdot 10^{-7}$. The reason for such low learning rates was the batch-based training implemented on GPU via OpenCL to achieve higher parallelism and reduce computation time. In this implementation, $\Delta c_f^k$ are summed for every event, causing large centroid and threshold updates, which are compensated by lowering the learning rate. Table \ref{tab:Sup3r-NMNIST} shows the result of this comparison.

\begin{table}
\centering
\caption{Sup3r NMNIST comparison. *HOTS implementation consisting of two hidden layers (32-96 units) of double exponential time surfaces and a SVM polynomial classifier \cite{rasetto2023building}. Sup3r and ANN consisting of a single hidden layer with 32 units.}
\begin{tabular}{l|lll}
            &Sup3r on N-MNIST& ANN on MNIST &HOTS on N-MNIST{\cite{rasetto2023building},*}\\ \hline
Test Accuracy&94.98\%& 95.61\% &91.27\%\\
 \% Processed Events&72.61\%&N/A &100\%\\ \end{tabular}
 
\label{tab:Sup3r-NMNIST}
\end{table}

Sup3r accuracy is comparable to an ANN on MNIST and higher than HOTS best results \cite{rasetto2023building} on a two-layered network (32-96 centroids) with double exponential decay time surfaces and a polynomial SVM classifier, while cutting off 28$\%$ of events.
While these results are promising, it is important to underline that they are preliminary. The reason is that this test was implemented on OpenCL for GPU acceleration prioritizing code readability and debugging rather than hardware optimization. This limited the number of experiments we run on N-MNIST. A single epoch of the N-MNIST dataset runs between 2 hours and 45 minutes on an NVIDIA 2060 Super, depending on the number of processed events in the network, as computation is done event by event. This is due to the high CPU usage to queue multiple kernels per event, the much bigger N-MNIST dataset compared to his frame-based counterpart (several thousands of sparse events compared to a single 28x28 frame),  and to HOTS, which requires several small matrix operations like multiplications and accumulation to process a single event. Exploiting vendor-specific optimized kernels, sparse libraries, kernel launching from GPU, and matrix operations should significantly reduce computation time and allow for an exhaustive hyperparameter analysis and testing on more complex architectures and datasets.

\section*{Conclusion}

In this paper, we presented Sup3r, a novel learning algorithm for training Hierarchy of Time Surfaces algorithms. Our results show that this algorithm can extract class-relevant features, reduce the number of processed events, and enable incremental and continual learning without the need for external classifiers. Moreover, preliminary results show that this algorithm could reach comparable accuracy to backpropagation on a small network. All the results together make Sup3r the best way to train HOTS networks end-to-end. Future works will have to focus on accelerating Sup3r on GPUs, testing it on more complex datasets and deep-layered architectures, and reducing the excessive number of hyperparameters. Another point of interest could be to adapt Sup3r to work on much more popular Spiking Neural Network models like Integrate and Fire neurons.

\bibliographystyle{unsrt}  
\bibliography{references}

\end{document}